\documentclass{article}

\usepackage[preprint]{nips_2018}

\usepackage[utf8]{inputenc} 
\usepackage[T1]{fontenc}    
\usepackage{hyperref}       
\usepackage{url}            
\usepackage{booktabs}       
\usepackage{amsfonts}       
\usepackage{nicefrac}       
\usepackage{microtype}      

\usepackage{amsmath}
\usepackage{amsthm}
\usepackage{algpseudocode,algorithm,algorithmicx}
\usepackage{graphicx}
\graphicspath{ {images/} }
\usepackage[usenames, dvipsnames]{color}

\usepackage{mathtools}
\definecolor{dark-gray}{gray}{.35}
\definecolor{myorange}{RGB}{246, 164, 16}
\definecolor{mygreen}{RGB}{1, 100, 3}
\usepackage{footnote}
\makesavenoteenv{table}
\makesavenoteenv{tabular}
\usepackage{cleveref}[2012/02/15]
\crefformat{footnote}{#2\footnotemark[#1]#3}

\title{The relativistic discriminator: a key element missing from standard GAN}

\author{
  Alexia Jolicoeur-Martineau \\
  Lady Davis Institute\\
  Montreal, Canada\\
  \texttt{alexia.jolicoeur-martineau@mail.mcgill.ca} \\
}

\begin{document}

\maketitle

\begin{abstract}
	
	In standard generative adversarial network (SGAN), the discriminator $D$ estimates the probability that the input data is real. The generator $G$ is trained to increase the probability that fake data is real. We argue that it should also simultaneously decrease the probability that real data is real because 1) this would account for \textit{a priori} knowledge that half of the data in the mini-batch is fake, 2) this would be observed with divergence minimization, and 3) in optimal settings, SGAN would be equivalent to integral probability metric (IPM) GANs. 

	We show that this property can be induced by using a “relativistic discriminator” which estimate the probability that the given real data is more realistic than a randomly sampled fake data. We also present a variant in which the discriminator estimate the probability that the given real data is more realistic than fake data, on average. We generalize both approaches to non-standard GAN loss functions and we refer to them respectively as Relativistic GANs (RGANs) and Relativistic average GANs (RaGANs). We show that IPM-based GANs are a subset of RGANs which use the identity function.
	
	Empirically, we observe that 1) RGANs and RaGANs are significantly more stable and generate higher quality data samples than their non-relativistic counterparts, 2) Standard RaGAN with gradient penalty generate data of better quality than WGAN-GP while only requiring a single discriminator update per generator update (reducing the time taken for reaching the state-of-the-art by 400\%), and 3) RaGANs are able to generate plausible high resolutions images (256x256) from a very small sample (N=2011), while GAN and LSGAN cannot; these images are of significantly better quality than the ones generated by WGAN-GP and SGAN with spectral normalization. 

\end{abstract}

\section{Introduction}

Generative adversarial networks (GANs) \citep{hong2017generative} form a broad class of generative models in which a game is played between two competing neural networks, the discriminator $D$ and the generator $G$. $D$ is trained to discriminate real from fake data, while $G$ is trained to generate fake data that $D$ will mistakenly recognize as real. In the original GAN by \citet{GAN}, which we refer to as Standard GAN (SGAN), $D$ is a classifier, thus it is predicting the probability that the input data is real. When $D$ is optimal, the loss function of SGAN is approximately equal to the Jensen–Shannon divergence (JSD) \citep{GAN}. 

SGAN has two variants for the generator loss functions: saturating and non-saturating. In practice, the former has been found to be very unstable, while the latter has been found to more stable \citep{GAN}. Under certain conditions, \citet{GANTheorems} proved that, if real and fake data are perfectly classified, the saturating loss has zero gradient and the non-saturating loss has non-zero but volatile gradient. In practice, this mean that the discriminator in SGAN often cannot be trained to optimality or with a too high learning rate; otherwise, gradients may vanish and, if so, training will stop. This problem is generally more noticeable in high-dimensional setting (e.g., high resolution images and discriminator architectures with high expressive power) given that there are more degrees of freedom available to reach perfect classification of the training set.

To improve on SGAN, many GAN variants have been suggested using different loss functions and discriminators that are not classifiers (e.g., LSGAN \citep{LSGAN}, WGAN \citep{WGAN}). Although these approaches have partially succeeded in improving stability and data quality, the large-scale study by \citet{lucic2017gans} suggests that these approaches do not consistently improve on SGAN. Additionally, some of the most successful approaches, such as WGAN-GP \citep{WGAN-GP}, are much more computationally demanding than SGAN.

Many of the recent successful GANs variants have been based on Integral probability metrics (IPMs) \citep{muller1997integral} (e.g., WGAN \citep{WGAN}, WGAN-GP\citep{WGAN-GP}, Sobolev GAN \citep{mroueh2017sobolev}, Fisher GAN \citep{Fisher}). In IPM-based GANs, the discriminator is real-valued and constrained to a specific class of function so that it does not grow too quickly; this act as a form of regularization which prevents $D$ from becoming too strong (i.e., almost perfectly classifying real from fake data). In practice, we generally observe that the discriminator of IPM-based GANs can be trained for many iterations without causing vanishing gradients. 
 
IPM constraints have been shown to be similarly beneficial in non-IPM-based GANs. The constraint of WGAN (i.e., Lipschitz discriminator) has been shown to be beneficial in other GANs through spectral normalization \citep{miyato2018spectral}. The constraint of WGAN-GP (i.e., discriminator with gradient norm equal to 1 around real and fake data) has been shown to be beneficial in SGAN \citep{ManyPaths} (along with a very similar gradient penalty by \cite{DRAGAN}). Although this shows that certain IPM constraints improve the stability of GANs, it does not explain why IPMs generally provide increased stability over other metrics/divergences in GANs (e.g., JSD for SGAN, $f$-divergences for $f$-GANs \citep{F-GAN}).

In this paper, we argue that non-IPM-based GANs are missing a key ingredient, a relativistic discriminator, which IPM-based GANs already possess. We show that a relativistic discriminator is necessary to make GANs analogous to divergence minimization and produce sensible predictions based on the \textit{a priori} knowledge that half of the samples in the mini-batch are fake. We provide empirical evidence showing that GANs with a relativistic discriminator are more stable and produce data of higher quality.

\section{Background}

\subsection{Generative adversarial networks}

GANs can be defined very generally in terms of the discriminator in the following way:
\begin{equation}
L_D = \mathbb{E}_{x_r \sim \mathbb{P}}\left[ \tilde{f}_1(D(x_r)) \right] + \mathbb{E}_{z \sim \mathbb{P}_z} \left[ \tilde{f}_2(D(G(z))) \right],
\end{equation}
and
\begin{equation}
L_G = \mathbb{E}_{x_r \sim \mathbb{P}}\left[ \tilde{g}_1(D(x_r)) \right] + \mathbb{E}_{z \sim \mathbb{P}_z} \left[ \tilde{g}_2(D(G(z))) \right],
\end{equation}
where $\tilde{f}_1$, $\tilde{f}_2$, $\tilde{g}_1$, $\tilde{g}_2$ are scalar-to-scalar functions, $\mathbb{P}$ is the distribution of real data,  $\mathbb{P}_z$ is generally a multivariate normal distribution centered at $0$ with variance 1, $D(x)$ is the discriminator evaluated at $x$, $G(z)$ is the generator evaluated at z ($\mathbb{Q}$ is the distribution of fake data, thus of $G(z)$). Note that, through the paper, we refer to real data as $x_r$ and fake data as $x_f$. Without loss of generality, we assume that both $L_D$ and $L_G$ are loss functions to be minimized.

Most GANs can be separated into two classes: non-saturating and saturating loss functions. GANs with the saturating loss are such that $\tilde{g}_1$=$-\tilde{f}_1$ and $\tilde{g}_2$=$-\tilde{f}_2$, while GANs with the non-saturating loss are such that $\tilde{g}_1$=$\tilde{f}_2$ and $\tilde{g}_2$=$\tilde{f}_1$. Saturating GANs are most intuitive as they can be interpreted as alternating between maximizing and minimizing the same loss function. After training $D$ to optimality, the loss function is generally an approximation of a divergence (e.g., Jensen–Shannon divergence (JSD) for SGAN \citep{GAN}, $f$-divergences for F-GANs \citep{F-GAN}, and Wassertein distance for WGAN \citep{WGAN}). Thus, training $G$ to minimize $L_G$ can be roughly interpreted as minimizing the approximated divergence (although this is not technically true; see \citet{alexia2018beyonddivergence}). On the other hand, non-saturating GANs can be thought as optimizing the same loss function, but swapping real data with fake data (and vice-versa). In this article, unless otherwise specified, we assume a non-saturating loss for all GANs.

SGAN assumes a cross-entropy loss, i.e., $\tilde{f}_1(D(x))=-\log(D(x))$ and $\tilde{f}_2(D(x))=-\log(1-D(x))$, where $D(x)=\text{sigmoid}(C(x))$, and $C(x)$ is the non-transformed discriminator output (which we call the \textit{critic} as per \citet{WGAN}). In most GANs, $C(x)$ can be interpreted as \textit{how realistic the input data is}; a negative number means that the input data looks fake (e.g., in SGAN, $D(x)=\text{sigmoid}(-5)=0$), while a positive number means that the input data looks real (e.g., in SGAN, $D(x)=\text{sigmoid}(5)=1$).

In SGAN, the discriminator is said to output the probability that the input data is real. This is because minimizing the cross-entropy is equivalent to maximizing the log-likelihood of a Bernoulli variable. Thus, the output of $D$ is approximately Bernoulli distributed and representative of a probability.

\subsection{Integral probability metrics}

IPMs are statistical divergences represented mathematically as:
\[
IPM_{F} (\mathbb{P} || \mathbb{Q}) = \sup_{C \in \mathcal{F}} \mathbb{E}_{x \sim \mathbb{P}}[C(x)] - \mathbb{E}_{x \sim \mathbb{Q}}[C(x)],
\]
where $\mathcal{F}$ is a class of real-valued functions. 

IPM-based GANs can be defined using equation 1 and 2 assuming $\tilde{f}_1(D(x))=\tilde{g}_2(D(x))=-D(x)$ and $\tilde{f}_2(D(x))=\tilde{g}_1(D(x))=D(x)$, where $D(x)=C(x)$ (i.e., no transformation is applied). It can be observed that both discriminator and generator loss functions are unbounded and would diverge to $-\infty$ if optimized directly. However, IPMs assume that the discriminator is of a certain class of function that does not grow too quickly which prevent the loss functions from diverging. Each IPM applies a different constraint to the discriminator (e.g., WGAN assumes a Lipschitz $D$, WGAN-GP assumes that $D$ has gradient norm equal to 1 around real and fake data).

\section{Missing property of SGAN}

\subsection{Missing property}

We argue that the key missing property of SGAN is that the probability of real data being real ($D(x_{r})$) should decrease as the probability of fake data being real ($D(x_{f})$) increase. We provide three arguments suggesting that SGAN should have this property.

\subsection{Prior knowledge argument}

With adequate training, the discriminator is able to correctly classify most real samples as real and most fake samples as not real. Subsequently, after the generator is trained to "fool" the discriminator into thinking that fake samples are real samples, the discriminator classify most samples, real or fake, as real. This behavior is illogical considering the \textit{a priori} knowledge that half of the samples in the mini-batch are fake, as we explain below. 

After training the generator, given that both real and fake samples look equally real, the critic values ($C(x)$) of real and fake data may be very close, i.e., $C(x_f) \approx C(x_r)$ for most $x_r$ and $x_f$. Considering the fact that the discriminator is always shown half real data and half fake data, if the discriminator perceive all samples shown as equally real, it should assume that each sample has probability .50 of being real. However, in SGAN and other non-IPM-based GANs, we implicitly assume that the discriminator does not know that half the samples are fake. If the discriminator doesn't know, it could be possible that all samples shown are real. Thus, if all samples look real, it would be reasonable to assume that they are indeed all real ($D(x)\approx 1$ for all $x$).

Assuming that the generator is trained with a strong learning rate or for many iterations; in addition to both real and fake samples being classified as real, fake samples may appear to be more realistic than real samples, i.e., $C(x_f)>C(x_r)$ for most $x_r$ and $x_f$. In that case, considering that half of the samples are fake, the discriminator should assign a higher probability of being fake to real samples rather than classify all samples are real.

In summary, by not decreasing $D(x_{r})$ as $D(x_{f})$ increase, SGAN completely ignores the \textit{a priori} knowledge that half of the mini-batch samples are fake. Unless one makes the task of the discriminator more difficult (using regularization or lower learning rates), the discriminator does not make reasonable predictions. On the other hand, IPM-based GANs implicitly account for the fact that some of the samples must be fake because they compare how realistic real data is compared to fake data. This provides an intuitive argument to why the discriminator in SGAN (and GANs in general) should depends on both real and fake data.

\subsection{Divergence minimization argument}

In SGAN, we have that the discriminator loss function is equal to the Jensen–Shannon divergence (JSD) \citep{GAN}. Therefore, calculating the JSD can be represented as solving the following maximum problem:

\begin{equation}
JSD(\mathbb{P} || \mathbb{Q}) = \frac{1}{2} \left(log(4) + \max_{D:X \rightarrow [0,1]} \mathbb{E}_{x_r \sim \mathbb{P}}[\log(D(x_r))] + \mathbb{E}_{x_f \sim \mathbb{Q}}[log \left(1 - D(x_f) \right)] \right).
\end{equation}

The JSD is minimized ($JSD(\mathbb{P} || \mathbb{Q})=0$) when $D(x_r) = D(x_f) = \frac{1}{2}$ for all $x_r \in \mathbb{P}$ and $x_f \in \mathbb{Q}$ and maximized ($JSD(\mathbb{P} || \mathbb{Q})=\log(2)$) when $D(x_r)=1$, $D(x_f)=0$ for all $x_r \in \mathbb{P}$ and $x_f \in \mathbb{Q}$. Thus, if we were directly minimizing the divergence from maximum to minimum, we would expect $D(x_r)$ to smoothly decrease from 1 to .50 for most $x_r$ and $D(x_f)$ to smoothly increase from 0 to .50 for most $x_f$ (Figure 1a). However, when minimizing the saturating loss in SGAN, we are only increasing $D(x_f)$, we are not decreasing $D(x_r)$ (Figure 1b). Furthermore, we are bringing $D(x_f)$ closer to 1 rather than .50. This means that SGAN dynamics are very different from the minimization of the JSD. To bring SGAN closer to divergence minimization, training the generator should not only increase $D(x_f)$ but also decrease $D(x_r)$ (Figure 1c).

\begin{figure}
	\centering
	\includegraphics[scale=0.5]{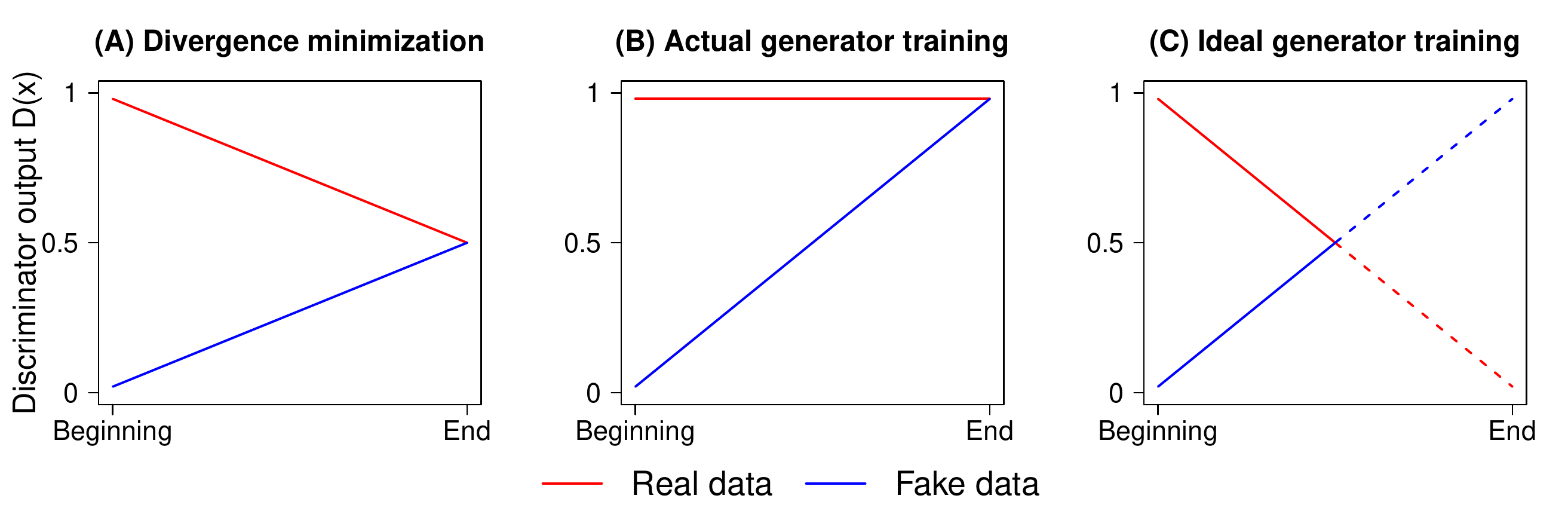}
	\caption{Expected discriminator output of the real and fake data for the a) direct minimization of the Jensen–Shannon divergence, b) actual training of the generator to minimize its loss function, and c) ideal training of the generator to minimize its loss function (lines are dotted when they cross beyond the equilibrium to signify that this may or may not be necessary).}
\end{figure}

\subsection{Gradient argument}

Let's compare the gradient steps of standard GAN and IPM-based GANs for further insight. It can be shown that the gradients of the discriminator and generator in non-saturating SGAN are respectively:
\begin{equation}
	\nabla_{w}L_D^{GAN} = -\mathbb{E}_{x_r \sim \mathbb{P}}\left[ (1-D(x_r)) \nabla_{w} C(x_r) \right] + \mathbb{E}_{x_f \sim \mathbb{Q_\theta}} \left[ D(x_f) \nabla_{w} C(x_f) \right],
\end{equation}
\begin{equation}
\nabla_{\theta}L_G^{GAN} = -\mathbb{E}_{z \sim \mathbb{P}_z} \left[ (1-D(G(z))) \nabla_{x} C(G(z)) J_{\theta} G(z) \right],
\end{equation}
where $J$ is the Jacobian.

It can be shown that the gradients of the discriminator and generator in IPM-based GANs are respectively:
\begin{equation}
\nabla_{w}L_D^{IPM} = - \mathbb{E}_{x_r \sim \mathbb{P}} [\nabla_{w}C(x_r)] + \mathbb{E}_{x_f \sim \mathbb{Q_\theta}}[\nabla_{w}C(x_f)],
\end{equation}
\begin{equation}
\nabla_{\theta}L_G^{IPM} = - \mathbb{E}_{z \sim \mathbb{P}_z}[\nabla_{x} C(G(z)) J_{\theta} G(z)],
\end{equation}
where $C(x) \in \mathcal{F}$ (the class of functions assigned by the IPM).

From these equations, it can be observed that SGAN leads to the same dynamics as IPM-based GANs when we have that:
\begin{enumerate}
	\item $D(x_r)=0$, $D(x_f)=1$ in the discriminator step of SGAN
	\item $D(x_f)=0$ in the generator step of SGAN.
	\item $C(x) \in \mathcal{F}$
\end{enumerate}
Assuming that the discriminator and generator are trained to optimality in each step and that it is possible to perfectly distinguish real from the fake data (strong assumption, but generally true early in training); we have that $D(x_r)=1$, $D(x_f)=0$ in the generator step and that $D(x_r)=1$, $D(x_f)=1$ in the discriminator step for most $x_r$ and $x_f$ (Figure 1b). Thus, the only missing assumption is that $D(x_r)=0$ in the discriminator step. 

This means that SGAN could be equivalent to IPM-based GANs, in certain situations, if the generator could indirectly influence $D(x_r)$. Considering that IPM-based GANs are generally more stable than SGAN, it would be reasonable to expect that making SGAN closer to IPM-based GANs could improve its stability. 

In IPMs, both real and fake data equally contribute to the gradient of the discriminator's loss function. However, in SGAN, if the discriminator reach optimality, the gradient completely ignores real data. This means that if $D(x_r)$ does not indirectly change when training the discriminator to reduce $D(x_f)$ (which might happens if real and fake data have different supports or if $D$ has a very large capacity), the discriminator will stop learning what it means for data to be "real" and training will focus entirely on fake data. In which case, fake samples will not become more realistic and training will get stuck. On the other hand, if $D(x_r)$ always decreases when $D(x_f)$ increases, real data will always be incorporated in the gradient of the discriminator loss function. In our experiments, we observe that GANs with this property are able to learn in very difficult settings whereas traditional GANs become stuck early in training.

\section{Method}

\subsection{Relativistic standard GAN}

In standard GAN, the discriminator can be defined, in term of the non-transformed layer $C(x)$, as $D(x)=\text{sigmoid}(C(x))$. A simple way to make discriminator relativistic (i.e., having the output of $D$ depends on both real and fake data) is to sample from real/fake data pairs $\tilde{x}=(x_r,x_f)$ and define it as $D(\tilde{x}) = \text{sigmoid}(C(x_r)-C(x_f))$. 

We can interpret this modification in the following way: \textbf{\textit{the discriminator estimates the probability that the given real data is more realistic than a randomly sampled fake data}}. Similarly, we can define $D_{rev}(\tilde{x}) = \text{sigmoid}(C(x_f)-C(x_r))$ as the probability that the given fake data is more realistic than a randomly sampled real data. An interesting property of this discriminator is that we do not need to include $D_{rev}$ in the loss function through $\log(1-D_{rev}(\tilde{x}))$ because we have that $1-D_{rev}(\tilde{x}) = 1 - \text{sigmoid}(C(x_f)-C(x_r)) = \text{sigmoid}(C(x_r)-C(x_f)) = D(\tilde{x})$; thus, $\log(D(\tilde{x}))$ = $\log(1-D_{rev}(\tilde{x}))$.

The discriminator and generator (non-saturating) loss functions of the Relativistic Standard GAN (RSGAN) can be written as:
\begin{equation}
L_D^{RSGAN} = -\mathbb{E}_{(x_r,x_f) \sim (\mathbb{P},\mathbb{Q})}\left[ \log (\text{sigmoid}(C(x_r)-C(x_f))) \right].
\end{equation}
\begin{equation}
L_G^{RSGAN} = -\mathbb{E}_{(x_r,x_f) \sim (\mathbb{P},\mathbb{Q})}\left[ \log (\text{sigmoid}(C(x_f)-C(x_r))) \right].
\end{equation}

\subsection{Relativistic GANs}

More generally, we consider any discriminator defined as $a(C(x_r)-C(x_f))$, where $a$ is the activation function, to be relativistic. This means that almost any GAN can have a relativistic discriminator. This forms a new class of models which we call Relativistic GANs (RGANs). 

Most GANs can be parametrized very generally in terms of the critic:
\begin{equation}
L_D^{GAN} = \mathbb{E}_{x_r \sim \mathbb{P}}\left[ f_1(C(x_r)) \right] + \mathbb{E}_{x_f \sim \mathbb{Q}} \left[ f_2(C(x_f)) \right]
\end{equation}
and
\begin{equation}
L_G^{GAN} = \mathbb{E}_{x_r \sim \mathbb{P}}\left[ g_1(C(x_r)) \right] + \mathbb{E}_{x_f \sim \mathbb{Q}}\left[ g_2(C(x_f)) \right],
\end{equation}
where $f_1$, $f_2$, $g_1$, $g_2$ are scalar-to-scalar functions. If we use a relativistic discriminator, these GANs now have the following form:
\begin{equation}
L_D^{RGAN} = \mathbb{E}_{(x_r,x_f) \sim (\mathbb{P},\mathbb{Q})}\left[ f_1(C(x_r)-C(x_f)) \right] + \mathbb{E}_{(x_r,x_f) \sim (\mathbb{P},\mathbb{Q})} \left[ f_2(C(x_f)-C(x_r)) \right]
\end{equation}
and
\begin{equation}
L_G^{RGAN} = \mathbb{E}_{(x_r,x_f) \sim (\mathbb{P},\mathbb{Q})}\left[ g_1(C(x_r)-C(x_f)) \right] + \mathbb{E}_{(x_r,x_f) \sim (\mathbb{P},\mathbb{Q})}\left[ g_2(C(x_f)-C(x_r)) \right].
\end{equation}
IPM-based GANs represent a special case of RGAN where $f_1(y)=g_2(y)=-y$ and $f_2(y)=g_1(y)=y$. Importantly, $g_1$ is normally ignored in GANs because its gradient is zero since the generator does not influence it. However, in RGANs, $g_1$ is influenced by fake data, thus by the generator. Therefore, $g_1$ generally has a non-zero gradient and needs to be specified in the generator loss. This means that in most RGANs (except in IPM-based GANs because they use the identity function), \textbf{\textit{the generator is trained to minimize the full loss function envisioned rather than only half of it}}.

The formulation of RGANs can be simplified when we have the following two properties: (1) $f_2(-y) = f_1(y)$ and (2) the generator assumes a non-saturating loss ($g_1(y)=f_2(y)$ and $g_2(y)=f_1(y)$). These two properties are observed in standard GAN, LSGAN using symmetric labels (e.g., -1 and 1), IPM-based GANs, etc. With these two properties, RGANs with non-saturating loss can be formulated simply as:
\begin{equation}
L_D^{RGAN*} = \mathbb{E}_{(x_r,x_f) \sim (\mathbb{P},\mathbb{Q})}\left[ f_1(C(x_r)-C(x_f)) \right]
\end{equation}
and
\begin{equation}
L_G^{RGAN*} = \mathbb{E}_{(x_r,x_f) \sim (\mathbb{P},\mathbb{Q})}\left[ f_1(C(x_f)-C(x_r)) \right].
\end{equation}
Algorithm 1 shows how to train RGANs of this form.
\begin{algorithm}  
	\caption{Training algorithm for non-saturating RGANs with symmetric loss functions
		\label{alg:1}}
	\begin{algorithmic}
		\Require{The number of $D$ iterations $n_{D}$ ($n_{D}=1$ unless one seeks to train $D$ to optimality), batch size $m$, and functions $f$ which determine the objective function of the discriminator ($f$ is $f_1$ from equation 10 assuming that $f_2(-y) = f_1(y)$, which is true for many GANs).}
		\While{$\theta$ has not converged}
		\For{$t = 1, \dots, n_{D}$}
		\State Sample $\{x^{(i)}\}_{i=1}^m \sim \mathbb{P}$
		\State Sample $\{z^{(i)}\}_{i=1}^m \sim \mathbb{P}_z$
		\State Update $w$ using SGD by ascending with $\nabla_{w} \frac{1}{m} \sum_{i=1}^{m} \left[ f(C_w(x^{(i)})-C_w(G_{\theta}(z^{(i)}))) \right]$
		\EndFor 
		\State Sample $\{x^{(i)}\}_{i=1}^m \sim \mathbb{P}$
		\State Sample $\{z^{(i)}\}_{i=1}^m \sim \mathbb{P}_z$
		\State Update $\theta$ using SGD by ascending with $\nabla_{\theta} \frac{1}{m} \sum_{i=1}^{m} \left[ f(C_w(G_{\theta}(z^{(i)}))-C_w(x^{(i)})) \right]$
		\EndWhile
	\end{algorithmic}
\end{algorithm}

\subsection{Relativistic average GANs}

Although the relative discriminator provide the missing property that we want in GANs (i.e., $G$ influencing $D(x_r)$), its interpretation is different from the standard discriminator. Rather than measuring “the probability that the input data is real”, it is now measuring “the probability that the input data is more realistic than a randomly sampled data of the opposing type (fake if the input is real or real if the input is fake)”. To make the relativistic discriminator act more globally, as in its original definition, our initial idea was to focus on the average of the relativistic discriminator over random samples of data of the opposing type. This can be conceptualized in the following way:
\begin{align*}
P(x_r \text{ is real}) &\vcentcolon= \mathbb{E}_{x_f \sim \mathbb{Q}}[P(x_r \text{ is more real than } x_f)] \\
&\phantom{\vcentcolon}= \mathbb{E}_{x_f \sim \mathbb{Q}}[\text{sigmoid}(C(x_r)-C(x_f))] \\
&\phantom{\vcentcolon}= \mathbb{E}_{x_f \sim \mathbb{Q}}[D(x_r, x_f)],
\end{align*}
\begin{align*}
P(x_f \text{ is real}) &\vcentcolon= \mathbb{E}_{x_r \sim \mathbb{P}}[P(x_f \text{ is more real than } x_r)] \\
&\phantom{\vcentcolon}= \mathbb{E}_{x_r \sim \mathbb{P}}[\text{sigmoid}(C(x_f)-C(x_r))] \\
&\phantom{\vcentcolon}= \mathbb{E}_{x_r \sim \mathbb{P}}[D(x_f, x_r)],
\end{align*}
where $D(x_r,x_f)=\text{sigmoid}(C(x_r)-C(x_f))$.

Then, the following loss function for $D$ could be applied:
\begin{equation}
L_D = -\mathbb{E}_{x_r \sim \mathbb{P}}\left[ \log \left( \mathbb{E}_{x_f \sim \mathbb{Q}}[D(x_r, x_f)] \right)) \right] - \mathbb{E}_{x_f \sim \mathbb{Q}} \left[ \log \left( 1 -  \mathbb{E}_{x_r \sim \mathbb{P}}[D(x_f, x_r)]\right) \right].
\end{equation}
The main problem with this idea is that it would require looking at all possible combinations of real and fake data in the mini-batch. This would transform the problem from $\mathcal{O}(m)$ to $\mathcal{O}(m^2)$ complexity, where $m$ is the batch size. This is problematic; therefore, we do not use this approach.

Instead, we propose to use the Relativistic average Discriminator (RaD) which compares the critic of the input data to the average critic of samples of the opposite type. The discriminator loss function for this approach can be formulated as: 
\begin{equation}
L_D^{RaSGAN} = -\mathbb{E}_{x_r \sim \mathbb{P}}\left[\log \left( \bar{D}(x_r) \right)) \right] - \mathbb{E}_{x_f \sim \mathbb{Q}} \left[\log \left(1- \bar{D}(x_f) \right) \right],
\end{equation}
where
\begin{equation}
\begin{aligned} 
\bar{D}(x)= 
\begin{cases}
\text{sigmoid}(C(x)-\mathbb{E}_{x_f \sim \mathbb{Q}} C(x_f))& \text{if } x \text{ is real}\\
\text{sigmoid}(C(x)-\mathbb{E}_{x_r \sim \mathbb{P}} C(x_r)) & \text{if } x \text{ is fake}.
\end{cases}
\end{aligned}
\end{equation}
RaD has a more similar interpretation to the standard discriminator than the relativistic discriminator. With RaD, \textbf{\textit{the discriminator estimates the probability that the given real data is more realistic than fake data, on average}}. This approach has $\mathcal{O}(m)$ complexity. Table 1 shows an intuitive and memeful visual representation of how this approach works. 

\begin{table}
	\centering
	\caption{A illustrative example of the discriminator's output in standard GAN as traditionally defined $(P(x_r \text{ is real})=\text{sigmoid}(C(x_r)))$ versus the Relativistic average Discriminator (RaD) $(P(x_r \text{ is real} | \overline{C(x_f)})=\text{sigmoid}(C(x_r)-\overline{C(x_f)}))$. Breads represent real images, while dogs represent fake images.} \label{tab:corgi}
	\begin{tabular}{ccc} 
		\toprule
		Scenario &  Absolute probability & Relative probability \\
		& (Standard GAN) & (Relativistic average Standard GAN) \\
		\midrule
		\begin{tabular}{@{}c@{}}Real image looks real \\ \textbf{and} \\ fake images look fake \end{tabular}  & 
		\begin{minipage}[c]{0.15\textwidth}
			\centering
			\includegraphics[width=\textwidth]{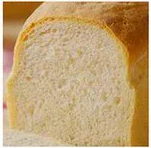}
		\end{minipage} & 
		\begin{minipage}[c]{0.30\textwidth}
			\centering
			\includegraphics[width=\textwidth]{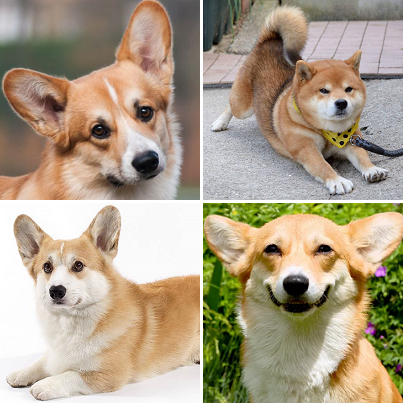}
		\end{minipage}
		\\
		& $C(x_r)=8$ & $\overline{C(x_f)}=-5$ \\
		& $P(x_r \text{ is bread})=1$ & $P(x_r \text{ is bread} | \overline{C(x_f)})=1$ \\
		\midrule
		\begin{tabular}{@{}c@{}}Real image looks real \\ \textbf{but} \\ fake images look \\ similarly real on average \end{tabular}  & 
		\begin{minipage}[c]{0.15\textwidth}
			\centering
			\includegraphics[width=\textwidth]{bread_real_1.png}
		\end{minipage} & 
		\begin{minipage}[c]{0.3\textwidth}
			\centering
			\includegraphics[width=\textwidth]{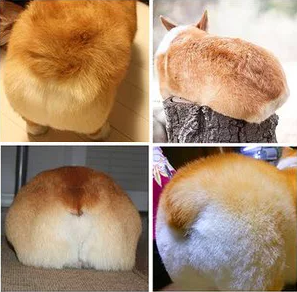}
		\end{minipage}
		\\
		& $C(x_r)=8$ & $\overline{C(x_f)}=7$ \\
		& $P(x_r \text{ is bread})=1$ & $P(x_r \text{ is bread} | \overline{C(x_f)})=.73$ \\
		\midrule
		\begin{tabular}{@{}c@{}}Real image looks fake \\ \textbf{but} \\ fake images look more \\ fake on average \end{tabular}  & 
		\begin{minipage}[c]{0.15\textwidth}
			\centering
			\includegraphics[width=\textwidth]{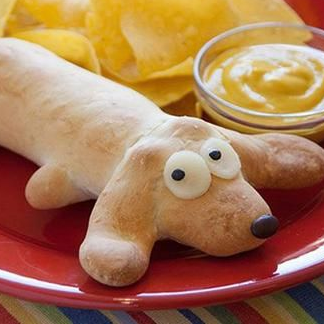}
		\end{minipage} & 
		\begin{minipage}[c]{0.30\textwidth}
			\centering
			\includegraphics[width=\textwidth]{corgi_fake_1.png}
		\end{minipage}
		\\
		& $C(x_r)=-3$ & $\overline{C(x_f)}=-5$ \\
		& $P(x_r \text{ is bread})=.05$ & $P(x_r \text{ is bread} | \overline{C(x_f)})=.88$ \\
		\bottomrule
	\end{tabular}
\end{table}

As before, we can generalize this approach to work with any GAN loss function using the following formulation:
\begin{equation}
L_D^{RaGAN} = \mathbb{E}_{x_r \sim \mathbb{P}}\left[ f_1\left( C(x_r)-\mathbb{E}_{x_f \sim \mathbb{Q}} C(x_f) \right)) \right] + \mathbb{E}_{x_f \sim \mathbb{Q}} \left[ f_2 \left( C(x_f)-\mathbb{E}_{x_r \sim \mathbb{P}} C(x_r) \right) \right].
\end{equation}
\begin{equation}
L_G^{RaGAN} = \mathbb{E}_{x_r \sim \mathbb{P}}\left[ g_1 \left( C(x_r)-\mathbb{E}_{x_f \sim \mathbb{Q}} C(x_f) \right)) \right] + \mathbb{E}_{x_f \sim \mathbb{Q}} \left[ g_2 \left( C(x_f)-\mathbb{E}_{x_r \sim \mathbb{P}} C(x_r) \right) \right].
\end{equation}
We call this general approach Relativistic average GAN (RaGAN). See Algorithm 2 for how to train non-saturating RaGANs.

\begin{algorithm}  
	\caption{Training algorithm for non-saturating RaGANs
		\label{alg:1}}
	\begin{algorithmic}
		\Require{The number of $D$ iterations $n_{D}$ ($n_{D}=1$ unless one seek to train $D$ to optimality), batch size $m$, and functions $f_1$ and $f_2$ which determine the objective function of the discriminator (see equation 10).}
		\While{$\theta$ has not converged}
		\For{$t = 1, \dots, n_{D}$}
		\State Sample $\{x^{(i)}\}_{i=1}^m \sim \mathbb{P}$
		\State Sample $\{z^{(i)}\}_{i=1}^m \sim \mathbb{P}_z$
		\State Let $\overline{C_w(x_r)} = \frac{1}{m} \sum_{i=1}^{m} C_w(x^{(i)})$
		\State Let $\overline{C_w(x_f)} = \frac{1}{m} \sum_{i=1}^{m} C_w(G_{\theta}(z^{(i)}))$
		\State Update $w$ using SGD by ascending with 
		\State \quad $\nabla_{w} \frac{1}{m} \sum_{i=1}^{m} \left[ f_1(C_w(x^{(i)})-\overline{C_w(x_f)}) +  f_2(C_w(G_{\theta}(z^{(i)}))-\overline{C_w(x_r)}) \right]$
		\EndFor 
		\State Sample $\{x^{(i)}\}_{i=1}^m \sim \mathbb{P}$
		\State Sample $\{z^{(i)}\}_{i=1}^m \sim \mathbb{P}_z$
		\State Let $\overline{C_w(x_r)} = \frac{1}{m} \sum_{i=1}^{m} C_w(x^{(i)})$
		\State Let $\overline{C_w(x_f)} = \frac{1}{m} \sum_{i=1}^{m} C_w(G_{\theta}(z^{(i)}))$
		\State Update $\theta$ using SGD by ascending with 
		\State \quad $\nabla_{\theta} \frac{1}{m} \sum_{i=1}^{m} \left[ f_1(C_w(G_{\theta}(z^{(i)}))-\overline{C_w(x_r)}) + f_2(C_w(x^{(i)})-\overline{C_w(x_f)}) \right]$
		\EndWhile
	\end{algorithmic}
\end{algorithm}

\section{Experiments}

Experiments were conducted on the CIFAR-10 dataset \citep{krizhevsky2009learning} and the CAT dataset \citep{cat}. Code was written in Pytorch \citep{pytorch} and models were trained using the Adam optimizer \citep{Adam} for 100K generator iterations with seed 1 (which shows that we did not fish for the best seed, instead, we selected the seed \textit{a priori}). We report the Fréchet Inception Distance (FID) \citep{heusel2017gans}, a measure that is generally better correlated with data quality than the Inception Distance \citep{tricks} \citep{borji2018pros}; lower FID means that the generated images are of better quality. 

For the models architectures, we used the standard CNN described by \citet{miyato2018spectral} on CIFAR-10 and a relatively standard DCGAN architecture \citep{DCGAN} on CAT (see Appendix). We also provide the source code required to replicate all analyses presented in this paper (See our repository: www.github.com/AlexiaJM/RelativisticGAN).

\subsection{Easy/stable experiments}

In these analyses, we compared standard GAN (SGAN), least-squares GAN (LSGAN), Wassertein GAN improved (WGAN-GP), Hinge-loss GAN (HingeGAN) \citep{miyato2018spectral}, Relativistic SGAN (RSGAN), Relativistic average SGAN (RaSGAN), Relativistic average LSGAN (RaLSGAN), and Relativistic average HingeGAN (RaHingeGAN) using the standard CNN architecture on stable setups (See Appendix for details on the loss functions used). Additionally, we tested RSGAN and RaSGAN with the same gradient-penalty as WGAN-GP (named RSGAN-GP and RaSGAN-GP respectively). 

We used the following two known stable setups: (DCGAN setup) $lr=.0002$, $n_D=1$, $\beta_1=.50$ and $\beta_2=.999$ \citep{DCGAN}, and (WGAN-GP setup) $lr=.0001$, $n_D=5$, $\beta_1=.50$ and $\beta_2=.9$ \citep{WGAN-GP}, where $lr$ is the learning rate, $n_D$ is the number of discriminator updates per generator update, and $\beta_1$, $\beta_2$ are the ADAM momentum parameters. For optimal stability, we used batch norm \citep{BatchNorm} in $G$ and spectral norm \citep{miyato2018spectral} in $D$.

Results are presented in Table 2. We observe that RSGAN and RaSGAN generally performed better than SGAN. Similarly, RaHingeGAN performed better than HingeGAN. RaLSGAN performed on par with LSGAN, albeit sightly worse. WGAN-GP performed poorly in the DCGAN setup, but very well in the WGAN-GP setup. RasGAN-GP performed poorly; however, RSGAN-GP performed better than all other loss functions using only one discriminator update per generator update. Importantly, the resulting FID of 25.60 is on par with the lowest FID obtained for this architecture using spectral normalization, as reported by \citet{miyato2018spectral} (25.5). Overall, these results show that using a relativistic discriminator generally improve data generation quality and that RSGAN works very well in conjunction with gradient penalty to obtain state-of-the-art results.

\begin{table}
	\caption{Fréchet Inception Distance (FID) at exactly 100k generator iterations on the CIFAR-10 dataset using stable setups with different GAN loss functions. We used spectral norm in $D$ and batch norm in $G$. All models were trained using the same \textit{a priori} selected seed (seed=1).}
	\label{CIFAR10}
	\centering
	\begin{tabular}{ccc}
		\toprule
		& $lr=.0002$ & $lr=.0001$ \\
		& $\beta=(.50,.999)$ & $\beta=(.50,.9)$ \\
		Loss & $n_D=1$ & $n_D=5$ \\
		\cmidrule(){1-3}
		SGAN & 40.64 & 41.32 \\
		RSGAN & 36.61 & 55.29 \\
		RaSGAN & 31.98 &  37.92  \\
		\cmidrule(){1-3}
		LSGAN &  29.53 & 187.01 \\
		RaLSGAN &  30.92 & 219.39 \\
		\cmidrule(){1-3}
		HingeGAN & 49.53 & 80.85 \\
		RaHingeGAN & 39.12 &  37.72 \\
		\cmidrule(){1-3}
		WGAN-GP & 83.89 & \fontseries{b}\selectfont 27.81 \\
		RSGAN-GP & \fontseries{b}\selectfont 25.60 &  28.13 \\
		RaSGAN-GP & 331.86 &  \\
		\bottomrule
	\end{tabular}
\end{table}

\subsection{Hard /unstable experiments}

\subsection{CIFAR-10}

In these analyses, we compared SGAN, LSGAN, WGAN-GP, RSGAN, RaSGAN, RaLSGAN, and RaHingeGAN with the standard CNN architecture on unstable setups in CIFAR-10. Unless otherwise specified, we used $lr=.0002$, $\beta_1=.5$, $\beta_2=.999$, $n_D=1$, and batch norm \citep{BatchNorm} in $G$ and $D$. We tested the following four unstable setups: (1) $lr = .001$, (2) $\beta_1=.9$, $\beta_2=.9$, (3) no batch norm in $G$ or $D$, and (4) all activation functions replaced with Tanh in both $G$ and $D$ (except for the output activation function of $D$).

Results are presented in Table 3. We observe that RaLSGAN performed better than LSGAN in all setups. RaHingeGAN performed slightly worse than HingeGAN in most setups. RSGAN and RaSGAN performed better than SGAN in two out of four setups, although differences were small. WGAN-GP generally performed poorly which we suspect is due to the single discriminator update per generator update. Overall, this provide good support for the improved stability of using the relative discriminator with LSGAN, but not with HingeGAN and SGAN. Although results are worse for the relativistic discriminator in some settings, differences are minimal and probably reflect natural variations. 

It is surprising to observe low FID for SGAN without batch normalization considering its well-known difficulty with this setting \citep{WGAN}. Given these results, we suspected that CIFAR-10 may be too easy to fully observe the stabilizing effects of using the relative discriminator. Therefore, our next analyses were done on the more difficult CAT dataset with high resolution pictures.

\begin{table}
	\caption{Fréchet Inception Distance (FID) at exactly 100k generator iterations on the CIFAR-10 dataset using instable setups with different GAN loss functions. Unless otherwise specified, we used $lr=.0002$, $\beta=(.50,.999)$, $n_D=1$, and batch norm (BN) in $D$ and $G$. All models were trained using the same \textit{a priori} selected seed (seed=1).}
	\label{CIFAR10}
	\centering
	\begin{tabular}{ccccc}
		\toprule
		Loss & $lr=.001$ & $\beta=(.9,.9)$ & No BN & Tanh \\
		\cmidrule{1-5}
		SGAN & 154.20 & 35.29 &  35.54 & 59.17 \\
		RSGAN &  50.95 &  45.12 &  37.11 & 77.21 \\
		RaSGAN & 55.55 &  43.46 & 41.96 &  54.42 \\
		\cmidrule{1-5}
		LSGAN & 52.27 & 225.94 & 38.54 & 147.87 \\
		RaLSGAN & \fontseries{b}\selectfont 33.33 & 48.92 & \fontseries{b}\selectfont 34.66 &  53.07 \\
		\cmidrule{1-5}
		HingeGAN &  43.28 & \fontseries{b}\selectfont 33.47 & 34.21 & 58.51 \\
		RaHingeGAN & 51.05 & 42.78 & 43.75 & \fontseries{b}\selectfont 50.69 \\
		\cmidrule{1-5}
		WGAN-GP & 61.97 & 104.95 & 85.27 & 59.94 \\
		\bottomrule
	\end{tabular}
\end{table}

\subsection{CAT}

CAT is a dataset containing around 10k pictures of cats with annotations. We cropped the pictures to the faces of the cats using those annotations. After removing outliers (hidden faces, blurriness, etc.), the CAT dataset contained 9304 images $\ge$ 64x64, 6645 images $\ge$ 128x128, and 2011 images $\ge$ 256x256. Previous analyses \footnote{As reported on https://ajolicoeur.wordpress.com/cats.} showed that the CAT dataset is particularly difficult in high-dimensions; SGAN generally has vanishing/exploding gradients with 64x64 images and is unable to generate 128x128 images without using certain tricks (e.g., unequal learning rates, Lipschitz discriminator, gradient penalty, etc.); this makes this dataset perfect for testing the stability of different GAN loss functions. 

We trained different GAN loss functions on 64x64, 128x128, 256x256 images. For 256x256 images, we compared RaGANs to known stable approaches: SpectralSGAN (SGAN with spectral normalization in $D$) and WGAN-GP. Although some approaches were able to train on 256x256 images, they did so with significant mode collapse. To alleviate this problem, for 256x256 images, we packed the discriminator \citep{pacgan} (i.e., $D$ took a concatenated pair of images instead of a single image). We looked at the mimimum, maximum, mean and standard deviation (SD) of the FID at 20k, 30k, ..., 100k generator iterations; results are presented in Table 4.

\begin{table}
	\caption{Minimum (min), maximum (max), mean, and standard deviation (SD) of the Fréchet Inception Distance (FID) calculated at 20k, 30k \ldots , 100k generator iterations on the CAT dataset with different GAN loss functions. The hyperparameters used were $lr=.0002$, $\beta=(.50,.999)$, $n_D=1$, and batch norm (BN) in $D$ and $G$. All models were trained using the same \textit{a priori} selected seed (seed=1).}
	\label{CIFAR10}
	\centering
	\begin{tabular}{ccccc}
		\toprule
		Loss & Min & Max & Mean & SD \\
		\cmidrule{1-5}
		\multicolumn{5}{c}{64x64 images (N=9304)} \\
		SGAN & 16.56 & 310.56 & 52.54 & 96.81 \\
		RSGAN & 19.03 & 42.05 & 32.16 & 7.01 \\
		RaSGAN & 15.38 & 33.11 & 20.53 & 5.68 \\
		\cmidrule{1-5}
		LSGAN & 20.27 & 224.97 & 73.62 & 61.02 \\
		RaLSGAN & \fontseries{b}\selectfont 11.97 & \fontseries{b}\selectfont 19.29 & \fontseries{b}\selectfont 15.61 & 2.55 \\
		\cmidrule{1-5}
		HingeGAN & 17.60 & 50.94 & 32.23 & 14.44 \\
		RaHingeGAN & 14.62 & 27.31 & 20.29 & 3.96 \\
		\cmidrule{1-5}
		RSGAN-GP & 16.41 & 22.34 & 18.20 & 1.82 \\
		RaSGAN-GP & 17.32 & 22 & 19.58 & \fontseries{b}\selectfont 1.81 \\
		\cmidrule{1-5}
		\multicolumn{5}{c}{128x128 images (N=6645)} \\
		SGAN \footnote{\label{bad} Didn't converge, became stuck in the first few iterations.} & - & - & - & - \\
		RaSGAN & 21.05 & \fontseries{b}\selectfont 39.65 & 28.53 & \fontseries{b}\selectfont 6.52 \\
		\cmidrule{1-5}
		LSGAN & 19.03 & 51.36 & 30.28 & 10.16 \\
		RaLSGAN & \fontseries{b}\selectfont 15.85 & 40.26 & \fontseries{b}\selectfont 22.36 & 7.53 \\
		\cmidrule{1-5}
		\multicolumn{5}{c}{256x256 images (N=2011)} \\
		SGAN\cref{bad} & - & - & - & - \\
		RaSGAN & \fontseries{b}\selectfont 32.11 & 102.76 & \fontseries{b}\selectfont 56.64 & 21.03 \\
		SpectralSGAN & 54.08 & \fontseries{b}\selectfont 90.43 & 64.92 & \fontseries{b}\selectfont 12.00 \\
		\cmidrule{1-5}
		LSGAN\cref{bad} & - & - & - & - \\
		RaLSGAN & 35.21 & 299.52 & 70.44 & 86.01 \\
		\cmidrule{1-5}
		WGAN-GP & 155.46 & 437.48 & 341.91 & 101.11 \\
		\bottomrule
	\end{tabular}
\end{table}

Overall, we observe lower minimum FID, maximum FID, mean and standard deviation (sd) for RGANs and RaGANs than their non-relativistic counterparts (SGAN, LSGAN, RaLSGAN).

In 64x64 resolution, both SGAN and LSGAN generated images with low FID, but they did so in a very unstable matter. For example, SGAN went from a FID of 17.50 at 30k iterations, to 310.56 at 40k iterations, and back to 27.72 at 50k iterations. Similarly, LSGAN went from a FID of 20.27 at 20k iterations, to 224.97 at 30k iterations, and back to 51.98 at 40k iterations. On the other hand, RaGANs were much more stable (lower max and SD) while also resulting in lower minimum FID. Using gradient-penalty did not improve data quality; however, it reduced the SD lower than without gradient penalty, thus increasing stability further.

SGAN was unable to converge on 128x128 or bigger images and LSGAN was unable to converge on 256x256 images. Meanwhile, RaGANs were able to generate plausible images with low FID in all resolutions. Although SpectralSGAN and WGAN-GP were able to generate 256x256 images of cats, the samples they generated were of poor quality (high FID). Thus, in this very difficult setting, relativism provided a greater improvement in quality than gradient penalty or spectral normalization.

\section{Conclusion and future work}

In this paper, we proposed the relativistic discriminator as a way to fix and improve on standard GAN. We further generalized this approach to any GAN loss and introduced a generally more stable variant called RaD. Our results suggest that relativism significantly improve data quality and stability of GANs at no computational cost. Furthermore, using a relativistic discriminator with other tools of the trade (spectral norm, gradient penalty, etc.) may lead to better state-of-the-art. 

Future research is needed to fully understand the mathematical implications of adding relativism to GANs. Furthermore, our experiments were limited to certain loss functions using only one seed, due to computational constraints. More experiments are required to determine which relativistic GAN loss function is best over a wide-range of datasets and hyperparameters. We greatly encourage researchers and machine learning enthusiasts with greater computing power to experiment further with our approach.

\bibliographystyle{unsrtnat}

\appendix
\section*{Appendices}
\addcontentsline{toc}{section}{Appendices}
\renewcommand{\thesubsection}{\Alph{subsection}}

\subsection{Gradient step}

\subsubsection{SGAN}

\begin{align*}
\nabla_{w}L_D^{GAN} &= -\nabla_{w} \mathbb{E}_{x_r \sim \mathbb{P}}\left[ \log D(x_r) \right] - \nabla_{w} \mathbb{E}_{x_f \sim \mathbb{Q_\theta}}\left[ \log (1-D(x_f)) \right] \\
&= - \nabla_{w} \mathbb{E}_{x_r \sim \mathbb{P}}\left[ \log \left( \frac{e^{C(x_r)}}{e^{C(x_r)}+1} \right) \right] - \nabla_{w} \mathbb{E}_{x_f \sim \mathbb{Q_\theta}}\left[ \log \left(1 - \frac{e^{C(x_f)}}{e^{C(x_f)}+1} \right) \right] \\
&= -\nabla_{w} \mathbb{E}_{x_r \sim \mathbb{P}}\left[ C(x_r) - \log \left( e^{C(x_r)}+1 \right) \right] - \nabla_{w} \mathbb{E}_{x_f \sim \mathbb{Q_\theta}}\left[ \log(1) - \log \left( e^{C(x_f)}+1 \right) \right] \\
&= -\mathbb{E}_{x_r \sim \mathbb{P}}\left[ \nabla_{w} C(x_r)\right] + \mathbb{E}_{x_r \sim \mathbb{P}}\left[ \frac{e^{C(x_r)}}{e^{C(x_r)}+1} \nabla_{w} C(x_r) \right] +  \mathbb{E}_{x_f \sim \mathbb{Q_\theta}} \left[ \frac{e^{C(x_f)}}{e^{C(x_f)}+1} \nabla_{w} C(x_f) \right] \\
&= -\mathbb{E}_{x_r \sim \mathbb{P}}\left[ \nabla_{w} C(x_r)\right] + \mathbb{E}_{x_r \sim \mathbb{P}}\left[ D(x_r) \nabla_{w} C(x_r) \right] +  \mathbb{E}_{x_f \sim \mathbb{Q_\theta}} \left[ D(x_f) \nabla_{w} C(x_f) \right] \\
&= -\mathbb{E}_{x_r \sim \mathbb{P}}\left[ (1-D(x_r)) \nabla_{w} C(x_r) \right] +  \mathbb{E}_{x_f \sim \mathbb{Q_\theta}} \left[ D(x_f) \nabla_{w} C(x_f) \right]
\end{align*}

\begin{align*}
\nabla_{\theta}L_G^{GAN} &= -\nabla_{\theta} \mathbb{E}_{z \sim \mathbb{P}_z}\left[ \log D(G(z)) \right] \\
&= -\nabla_{\theta} \mathbb{E}_{z \sim \mathbb{P}_z} \left[ \log \left(\frac{e^{C(G(z))}}{e^{C(G(z))}+1} \right) \right] \\
&= -\nabla_{\theta} \mathbb{E}_{z \sim \mathbb{P}_z} \left[ C(G(z)) - \log \left( e^{C(G(z))}+1 \right) \right] \\
&= -\mathbb{E}_{z \sim \mathbb{P}_z} \left[ \nabla_{x} C(G(z)) J_{\theta} G(z) - \left( \frac{e^{C(G(z))}}{e^{C(G(z))}+1} \right) \nabla_{x} C(G(z)) J_{\theta} G(z) \right] \\
&= -\mathbb{E}_{z \sim \mathbb{P}_z} \left[ (1-D(G(z))) \nabla_{x} C(G(z)) J_{\theta} G(z) \right]
\end{align*}

\subsubsection{IPM-based GANs}

\begin{align*}
\nabla_{w}L_D^{IPM} &= -\nabla_{w} \mathbb{E}_{x_r \sim \mathbb{P}} [C(x_r)] + \nabla_{w} \mathbb{E}_{x_f \sim \mathbb{Q_\theta}}[C(x_f)] \\
&= -\mathbb{E}_{x_r \sim \mathbb{P}} [\nabla_{w}C(x_r)] + \mathbb{E}_{x_f \sim \mathbb{Q_\theta}}[\nabla_{w}C(x_f)]
\end{align*}

\begin{align*}
\nabla_{\theta}L_G^{IPM} &= - \nabla_{\theta} \mathbb{E}_{z \sim \mathbb{P}_z}[C(G(z))] \\
&= - \mathbb{E}_{z \sim \mathbb{P}_z}[\nabla_{x} C(G(z)) J_{\theta} G(z)]
\end{align*}

\subsection{Simplified form of relativistic saturating and non-saturating GANs}

Assuming $f_2(-y)=f_1(y)$, we have that
\begin{align*}
L_D^{RGAN} &= \mathbb{E}_{(x_r,x_f) \sim (\mathbb{P},\mathbb{Q})}\left[ f_1(C(x_r)-C(x_f)) \right] + \mathbb{E}_{(x_r,x_f) \sim (\mathbb{P},\mathbb{Q})} \left[ f_2(C(x_f)-C(x_r)) \right] \\
&= \mathbb{E}_{(x_r,x_f) \sim (\mathbb{P},\mathbb{Q})}\left[ f_1(C(x_r)-C(x_f)) \right] + \mathbb{E}_{(x_r,x_f) \sim (\mathbb{P},\mathbb{Q})} \left[ f_1(C(x_r)-C(x_f)) \right] \\
&= 2\mathbb{E}_{(x_r,x_f) \sim (\mathbb{P},\mathbb{Q})}\left[ f_1(C(x_r)-C(x_f)) \right].
\end{align*}

If $g_1(y)=-f_1(y)$ and $g_2(y)=-f_2(y)$ (saturating GAN), we have that
\begin{align*}
L_G^{RGAN-S} &= \mathbb{E}_{(x_r,x_f) \sim (\mathbb{P},\mathbb{Q})}\left[ g_1(C(x_r)-C(x_f)) \right] + \mathbb{E}_{(x_r,x_f) \sim (\mathbb{P},\mathbb{Q})}\left[ g_2(C(x_f)-C(x_r)) \right] \\
&= -\mathbb{E}_{(x_r,x_f) \sim (\mathbb{P},\mathbb{Q})}\left[f_1(C(x_r)-C(x_f)) \right] - \mathbb{E}_{(x_r,x_f) \sim (\mathbb{P},\mathbb{Q})}\left[f_2(C(x_f)-C(x_r)) \right] \\
&= -\mathbb{E}_{(x_r,x_f) \sim (\mathbb{P},\mathbb{Q})}\left[f_1(C(x_r)-C(x_f)) \right] - \mathbb{E}_{(x_r,x_f) \sim (\mathbb{P},\mathbb{Q})}\left[f_1(C(x_r)-C(x_f)) \right] \\
&= -2\mathbb{E}_{(x_r,x_f) \sim (\mathbb{P},\mathbb{Q})}\left[f_1(C(x_r)-C(x_f)) \right].
\end{align*}

If $g_1(y)=f_2(y)$ and $g_2(y)=f_1(y)$ (non-saturating GAN), we have that
\begin{align*}
L_G^{RGAN-NS} &= \mathbb{E}_{(x_r,x_f) \sim (\mathbb{P},\mathbb{Q})}\left[ g_1(C(x_r)-C(x_f)) \right] + \mathbb{E}_{(x_r,x_f) \sim (\mathbb{P},\mathbb{Q})}\left[ g_2(C(x_f)-C(x_r)) \right] \\
&= \mathbb{E}_{(x_r,x_f) \sim (\mathbb{P},\mathbb{Q})}\left[f_2(C(x_r)-C(x_f)) \right] + \mathbb{E}_{(x_r,x_f) \sim (\mathbb{P},\mathbb{Q})}\left[f_1(C(x_f)-C(x_r)) \right] \\
&= \mathbb{E}_{(x_r,x_f) \sim (\mathbb{P},\mathbb{Q})}\left[f_1(C(x_f)-C(x_r)) \right] + \mathbb{E}_{(x_r,x_f) \sim (\mathbb{P},\mathbb{Q})}\left[f_1(C(x_f)-C(x_r)) \right] \\
&= 2\mathbb{E}_{(x_r,x_f) \sim (\mathbb{P},\mathbb{Q})}\left[f_1(C(x_f)-C(x_r)) \right].
\end{align*}

\subsection{Loss functions used in experiments}

\subsubsection{SGAN (non-saturating)}

\begin{equation}
L_D^{SGAN} = -\mathbb{E}_{x_r \sim \mathbb{P}}\left[ \log\left( \text{sigmoid}(C(x_r)) \right) \right] - \mathbb{E}_{x_f \sim \mathbb{Q}} \left[ \log \left( 1-\text{sigmoid}(C(x_f)) \right) \right]
\end{equation} 
\begin{equation}
L_G^{SGAN} = -\mathbb{E}_{x_f \sim \mathbb{Q}} \left[ \log \left(\text{sigmoid}(C(x_f)) \right) \right]
\end{equation}

\subsubsection{RSGAN}

\begin{equation}
L_D^{RSGAN} = -\mathbb{E}_{(x_r,x_f) \sim (\mathbb{P},\mathbb{Q})}\left[ \log (\text{sigmoid}(C(x_r)-C(x_f))) \right]
\end{equation}
\begin{equation}
L_G^{RSGAN} = -\mathbb{E}_{(x_r,x_f) \sim (\mathbb{P},\mathbb{Q})}\left[ \log (\text{sigmoid}(C(x_f)-C(x_r))) \right]
\end{equation}

\subsubsection{RaSGAN}

\begin{equation}
L_D^{RaSGAN} = -\mathbb{E}_{x_r \sim \mathbb{P}}\left[ \log\left( \tilde{D}(x_r) \right) \right] - \mathbb{E}_{x_f \sim \mathbb{Q}} \left[ \log \left( 1 - \tilde{D}(x_f) \right) \right]
\end{equation} 
\begin{equation}
L_G^{RaSGAN} = -\mathbb{E}_{x_f \sim \mathbb{Q}}\left[ \log\left( \tilde{D}(x_f) \right) \right] - \mathbb{E}_{x_r \sim \mathbb{P}} \left[ \log \left( 1 - \tilde{D}(x_r) \right) \right]
\end{equation}
$\tilde{D}(x_r)=\text{sigmoid} \left( C(x_r)-\mathbb{E}_{x_f \sim \mathbb{Q}} C(x_f) \right)$ \\ $ \tilde{D}(x_f)=\text{sigmoid} \left( C(x_f)-\mathbb{E}_{x_r \sim \mathbb{P}} C(x_r) \right)$

\subsubsection{LSGAN}

\begin{equation}
L_D^{LSGAN} = \mathbb{E}_{x_r \sim \mathbb{P}}\left[  (C(x_r)-0)^2 \right] + \mathbb{E}_{x_f \sim \mathbb{Q}} \left[ (C(x_f)-1)^2 \right]
\end{equation} 
\begin{equation}
L_G^{LSGAN} = \mathbb{E}_{x_f \sim \mathbb{Q}} \left[ (C(x_f)-0)^2 \right]
\end{equation}

\subsubsection{RaLSGAN}

\begin{equation}
L_D^{RaLSGAN} = \mathbb{E}_{x_r \sim \mathbb{P}}\left[  (C(x_r)-\mathbb{E}_{x_f \sim \mathbb{Q}} C(x_f)-1)^2 \right] + \mathbb{E}_{x_f \sim \mathbb{Q}} \left[ (C(x_f)-\mathbb{E}_{x_r \sim \mathbb{P}} C(x_r)+1)^2 \right]
\end{equation} 
\begin{equation}
L_G^{RaLSGAN} = \mathbb{E}_{x_f \sim \mathbb{P}}\left[  (C(x_f)-\mathbb{E}_{x_r \sim \mathbb{P}} C(x_r)-1)^2 \right] + \mathbb{E}_{x_r \sim \mathbb{P}} \left[ (C(x_r)-\mathbb{E}_{x_f \sim \mathbb{Q}} C(x_f)+1)^2 \right]
\end{equation}

\subsubsection{HingeGAN}

\begin{equation}
L_D^{HingeGAN} = \mathbb{E}_{x_r \sim \mathbb{P}}\left[ \max(0, 1-C(x_r)) \right] + \mathbb{E}_{x_f \sim \mathbb{Q}} \left[ \max(0, 1+C(x_f)) \right]
\end{equation} 
\begin{equation}
L_G^{HingeGAN} = -\mathbb{E}_{x_f \sim \mathbb{Q}} \left[ C(x_f) \right]
\end{equation}

\subsubsection{RaHingeGAN}

\begin{equation}
L_D^{HingeGAN} = \mathbb{E}_{x_r \sim \mathbb{P}}\left[ \max(0, 1-\tilde{D}(x_r)) \right] + \mathbb{E}_{x_f \sim \mathbb{Q}} \left[ \max(0, 1+\tilde{D}(x_f)) \right]
\end{equation} 
\begin{equation}
L_G^{HingeGAN} = \mathbb{E}_{x_f \sim \mathbb{P}}\left[ \max(0, 1-\tilde{D}(x_f)) \right] + \mathbb{E}_{x_r \sim \mathbb{Q}} \left[ \max(0, 1+\tilde{D}(x_r)) \right]
\end{equation}
$\tilde{D}(x_r) = C(x_r)-\mathbb{E}_{x_f \sim \mathbb{Q}} C(x_f)$ \\
$\tilde{D}(x_f) = C(x_f)-\mathbb{E}_{x_r \sim \mathbb{P}} C(x_r)$

\subsubsection{WGAN-GP}

\begin{equation}
L_D^{WGAN-GP} = -\mathbb{E}_{x_r \sim \mathbb{P}}\left[ C(x_r) \right] + \mathbb{E}_{x_f \sim \mathbb{Q}} \left[ C(x_f) \right] + \lambda \mathbb{E}_{\hat{x} \sim \mathbb{P}_{\hat{x}}} \left[ \left( || \nabla_{\hat{x}} C(\hat{x}) ||_2 - 1 \right)^2 \right]
\end{equation} 
\begin{equation}
L_G^{WGAN-GP} = -\mathbb{E}_{x_f \sim \mathbb{Q}} \left[ C(x_f) \right]
\end{equation}
$\mathbb{P}_{\hat{x}}$ is the distribution of $\hat{x} = \epsilon x_r + (1-\epsilon) x_f$, where $x_r \sim \mathbb{P}$, $x_f \sim \mathbb{Q}$, $\epsilon \sim U[0,1]$.

\subsubsection{RSGAN-GP}

\begin{equation}
L_D^{RSGAN} = -\mathbb{E}_{(x_r,x_f) \sim (\mathbb{P},\mathbb{Q})}\left[ \log (\text{sigmoid}(C(x_r)-C(x_f))) \right] + \lambda \mathbb{E}_{\hat{x} \sim \mathbb{P}_{\hat{x}}} \left[ \left( || \nabla_{\hat{x}} C(\hat{x}) ||_2 - 1 \right)^2 \right]
\end{equation}
\begin{equation}
L_G^{RSGAN} = -\mathbb{E}_{(x_r,x_f) \sim (\mathbb{P},\mathbb{Q})}\left[ \log (\text{sigmoid}(C(x_f)-C(x_r))) \right]
\end{equation}
$\mathbb{P}_{\hat{x}}$ is the distribution of $\hat{x} = \epsilon x_r + (1-\epsilon) x_f$, where $x_r \sim \mathbb{P}$, $x_f \sim \mathbb{Q}$, $\epsilon \sim U[0,1]$.

\subsubsection{RaSGAN-GP}

\begin{equation}
L_D^{RaSGAN} = -\mathbb{E}_{x_r \sim \mathbb{P}}\left[ \log\left( \tilde{D}(x_r) \right) \right] - \mathbb{E}_{x_f \sim \mathbb{Q}} \left[ \log \left( 1 - \tilde{D}(x_f) \right) \right] + \lambda \mathbb{E}_{\hat{x} \sim \mathbb{P}_{\hat{x}}} \left[ \left( || \nabla_{\hat{x}} C(\hat{x}) ||_2 - 1 \right)^2 \right]
\end{equation} 
\begin{equation}
L_G^{RaSGAN} = -\mathbb{E}_{x_f \sim \mathbb{Q}}\left[ \log\left( \tilde{D}(x_f) \right) \right] - \mathbb{E}_{x_r \sim \mathbb{P}} \left[ \log \left( 1 - \tilde{D}(x_r) \right) \right]
\end{equation}
$\tilde{D}(x_r)=\text{sigmoid} \left( C(x_r)-\mathbb{E}_{x_f \sim \mathbb{Q}} C(x_f) \right)$ \\ $ \tilde{D}(x_f)=\text{sigmoid} \left( C(x_f)-\mathbb{E}_{x_r \sim \mathbb{P}} C(x_r) \right)$ \\
$\mathbb{P}_{\hat{x}}$ is the distribution of $\hat{x} = \epsilon x_r + (1-\epsilon) x_f$, where $x_r \sim \mathbb{P}$, $x_f \sim \mathbb{Q}$, $\epsilon \sim U[0,1]$.

\subsection{Architectures}

\subsubsection{Standard CNN}

\begin{tabular}{c}
	Generator \\
	\toprule\midrule
	$z \in \mathbb{R}^{128} \sim N(0,I)$ \\
	\midrule
	linear, 128 -> 512*4*4 \\
	\midrule
	Reshape, 512*4*4 -> 512 x 4 x 4 \\
	\midrule
	ConvTranspose2d 4x4, stride 2, pad 1, 512->256 \\
	\midrule
	BN and ReLU \\
	\midrule
	ConvTranspose2d 4x4, stride 2, pad 1, 256->128 \\
	\midrule
	BN and ReLU \\
	\midrule
	ConvTranspose2d 4x4, stride 2, pad 1, 128->64 \\
	\midrule
	BN and ReLU \\
	\midrule
	ConvTranspose2d 3x3, stride 1, pad 1, 64->3 \\
	\midrule
	Tanh \\
	\bottomrule
\end{tabular} 
\quad
\begin{tabular}{c}
	Discriminator \\
	\toprule\midrule
	$x \in \mathbb{R}^{\text{3x32x32}}$ \\
	\midrule
	Conv2d 3x3, stride 1, pad 1, 3->64 \\
	\midrule
	LeakyReLU 0.1 \\
	\midrule
	Conv2d 4x4, stride 2, pad 1, 64->64 \\
	\midrule
	LeakyReLU 0.1 \\
	\midrule
	Conv2d 3x3, stride 1, pad 1, 64->128 \\
	\midrule
	LeakyReLU 0.1 \\
	\midrule
	Conv2d 4x4, stride 2, pad 1, 128->128 \\
	\midrule
	LeakyReLU 0.1 \\
	\midrule
	Conv2d 3x3, stride 1, pad 1, 128->256 \\
	\midrule
	LeakyReLU 0.1 \\
	\midrule
	Conv2d 4x4, stride 2, pad 1, 256->256 \\
	\midrule
	LeakyReLU 0.1 \\
	\midrule
	Conv2d 3x3, stride 1, pad 1, 256->512 \\
	\midrule
	Reshape, 512 x 4 x 4 -> 512*4*4 \\
	\midrule
	linear, 512*4*4 -> 1 \\
	\bottomrule
\end{tabular}

\subsubsection{DCGAN 64x64}

\begin{tabular}{c}
	Generator \\
	\toprule\midrule
	$z \in \mathbb{R}^{128} \sim N(0,I)$ \\
	\midrule
	ConvTranspose2d 4x4, stride 1, pad 0, no bias, 128->512 \\
	\midrule
	BN and ReLU \\
	\midrule
	ConvTranspose2d 4x4, stride 2, pad 1, no bias, 512->256 \\
	\midrule
	BN and ReLU \\
	\midrule
	ConvTranspose2d 4x4, stride 2, pad 1, no bias, 256->128 \\
	\midrule
	BN and ReLU \\
	\midrule
	ConvTranspose2d 4x4, stride 2, pad 1, no bias, 128->64 \\
	\midrule
	BN and ReLU \\
	\midrule
	ConvTranspose2d 4x4, stride 2, pad 1, no bias, 64->3 \\
	\midrule
	Tanh \\
	\bottomrule
\end{tabular} 
\quad
\begin{tabular}{c}
	Discriminator \\
	\toprule\midrule
	$x \in \mathbb{R}^{\text{3x64x64}}$ \\
	\midrule
	Conv2d 4x4, stride 2, pad 1, no bias, 3->64 \\
	\midrule
	LeakyReLU 0.2 \\
	\midrule
	Conv2d 4x4, stride 2, pad 1, no bias, 64->128 \\
	\midrule
	BN and LeakyReLU 0.2 \\
	\midrule
	Conv2d 4x4, stride 2, pad 1, no bias, 128->256 \\
	\midrule
	BN and LeakyReLU 0.2 \\
	\midrule
	Conv2d 4x4, stride 2, pad 1, no bias, 256->512 \\
	\midrule
	BN and LeakyReLU 0.2 \\
	\midrule
	Conv2d 4x4, stride 2, pad 1, no bias, 512->1 \\
	\bottomrule
\end{tabular}

\subsubsection{DCGAN 128x128}

\begin{tabular}{c}
	Generator \\
	\toprule\midrule
	$z \in \mathbb{R}^{128} \sim N(0,I)$ \\
	\midrule
	ConvTranspose2d 4x4, stride 1, pad 0, no bias, 128->1024 \\
	\midrule
	BN and ReLU \\
	\midrule
	ConvTranspose2d 4x4, stride 2, pad 1, no bias, 1024->512 \\
	\midrule
	BN and ReLU \\
	\midrule
	ConvTranspose2d 4x4, stride 2, pad 1, no bias, 512->256 \\
	\midrule
	BN and ReLU \\
	\midrule
	ConvTranspose2d 4x4, stride 2, pad 1, no bias, 256->128 \\
	\midrule
	BN and ReLU \\
	\midrule
	ConvTranspose2d 4x4, stride 2, pad 1, no bias, 128->64 \\
	\midrule
	BN and ReLU \\
	\midrule
	ConvTranspose2d 4x4, stride 2, pad 1, no bias, 64->3 \\
	\midrule
	Tanh \\
	\bottomrule
\end{tabular} 
\quad
\begin{tabular}{c}
	Discriminator \\
	\toprule\midrule
	$x \in \mathbb{R}^{\text{3x128x128}}$ \\
	\midrule
	Conv2d 4x4, stride 2, pad 1, no bias, 3->64 \\
	\midrule
	LeakyReLU 0.2 \\
	\midrule
	Conv2d 4x4, stride 2, pad 1, no bias, 64->128 \\
	\midrule
	BN and LeakyReLU 0.2 \\
	\midrule
	Conv2d 4x4, stride 2, pad 1, no bias, 128->256 \\
	\midrule
	BN and LeakyReLU 0.2 \\
	\midrule
	Conv2d 4x4, stride 2, pad 1, no bias, 256->512 \\
	\midrule
	BN and LeakyReLU 0.2 \\
	\midrule
	Conv2d 4x4, stride 2, pad 1, no bias, 512->1024 \\
	\midrule
	BN and LeakyReLU 0.2 \\
	\midrule
	Conv2d 4x4, stride 2, pad 1, no bias, 1024->1 \\
	\bottomrule
\end{tabular}

\subsubsection{DCGAN 256x256}

\begin{tabular}{c}
	Generator \\
	\toprule\midrule
	$z \in \mathbb{R}^{128} \sim N(0,I)$ \\
	\midrule
	ConvTranspose2d 4x4, stride 1, pad 0, no bias, 128->1024 \\
	\midrule
	BN and ReLU \\
	\midrule
	ConvTranspose2d 4x4, stride 2, pad 1, no bias, 1024->512 \\
	\midrule
	BN and ReLU \\
	\midrule
	ConvTranspose2d 4x4, stride 2, pad 1, no bias, 512->256 \\
	\midrule
	BN and ReLU \\
	\midrule
	ConvTranspose2d 4x4, stride 2, pad 1, no bias, 256->128 \\
	\midrule
	BN and ReLU \\
	\midrule
	ConvTranspose2d 4x4, stride 2, pad 1, no bias, 128->64 \\
	\midrule
	BN and ReLU \\
	\midrule
	ConvTranspose2d 4x4, stride 2, pad 1, no bias, 64->32 \\
	\midrule
	BN and ReLU \\
	\midrule
	ConvTranspose2d 4x4, stride 2, pad 1, no bias, 64->3 \\
	\midrule
	Tanh \\
	\bottomrule
\end{tabular} 
\quad
\begin{tabular}{c}
	Discriminator (PACGAN2 \citep{pacgan}) \\
	\toprule\midrule
	$x_1 \in \mathbb{R}^{\text{3x256x256}}, x_2 \in \mathbb{R}^{\text{3x256x256}}$ \\
	\midrule
	Concatenate $[x_1, x_2] \in \mathbb{R}^{\text{6x256x256}}$ \\
	\midrule
	Conv2d 4x4, stride 2, pad 1, no bias, 6->32 \\
	\midrule
	LeakyReLU 0.2 \\
	\midrule
	Conv2d 4x4, stride 2, pad 1, no bias, 32->64 \\
	\midrule
	LeakyReLU 0.2 \\
	\midrule
	Conv2d 4x4, stride 2, pad 1, no bias, 64->128 \\
	\midrule
	BN and LeakyReLU 0.2 \\
	\midrule
	Conv2d 4x4, stride 2, pad 1, no bias, 128->256 \\
	\midrule
	BN and LeakyReLU 0.2 \\
	\midrule
	Conv2d 4x4, stride 2, pad 1, no bias, 256->512 \\
	\midrule
	BN and LeakyReLU 0.2 \\
	\midrule
	Conv2d 4x4, stride 2, pad 1, no bias, 512->1024 \\
	\midrule
	BN and LeakyReLU 0.2 \\
	\midrule
	Conv2d 4x4, stride 2, pad 1, no bias, 1024->1 \\
	\bottomrule
\end{tabular}

\subsection{Samples}

This shows a selection of cats from certain models. Images shown are from the lowest FID registered at every 10k generator iterations. Given space constraint, with higher resolutions cats, we show some of the nicer looking cats for each approach, there are evidently some worse looking cats \footnote{See https://github.com/AlexiaJM/RelativisticGAN/tree/master/images/full\_minibatch for all cats of the mini-batch.}.

\begin{figure}[H]
	\centering
	\includegraphics[width=390pt]{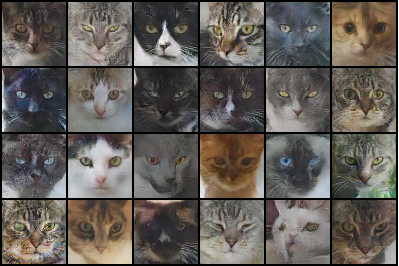}
	\caption{64x64 cats with RaLSGAN (FID = 11.97)}
\end{figure}

\begin{figure}[H]
	\centering
	\includegraphics[width=390pt]{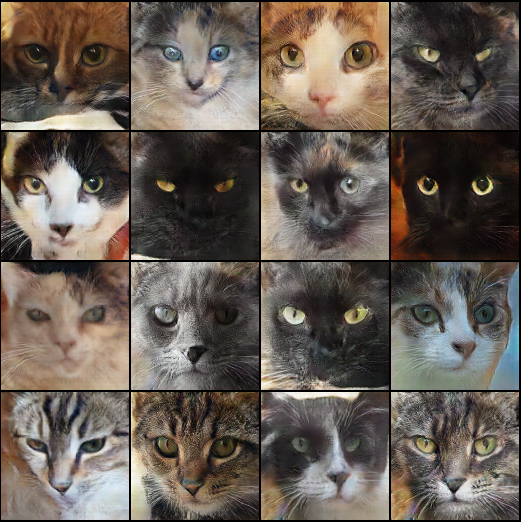}
	\caption{128x128 cats with RaLSGAN (FID = 15.85)}
\end{figure}

\begin{figure}[H]
	\centering
	\includegraphics[width=256pt]{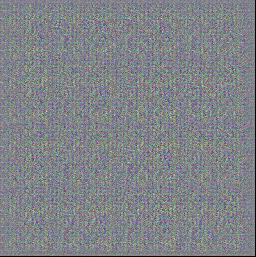}
	\caption{256x256 cats with GAN (5k iterations)}
\end{figure}

\begin{figure}[H]
	\centering
	\includegraphics[width=256pt]{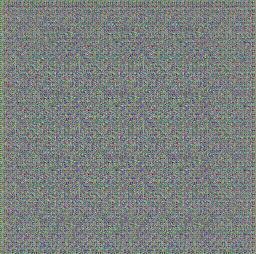}
	\caption{256x256 cats with LSGAN (5k iterations)}
\end{figure}

\begin{figure}[H]
	\centering
	\includegraphics[width=390pt]{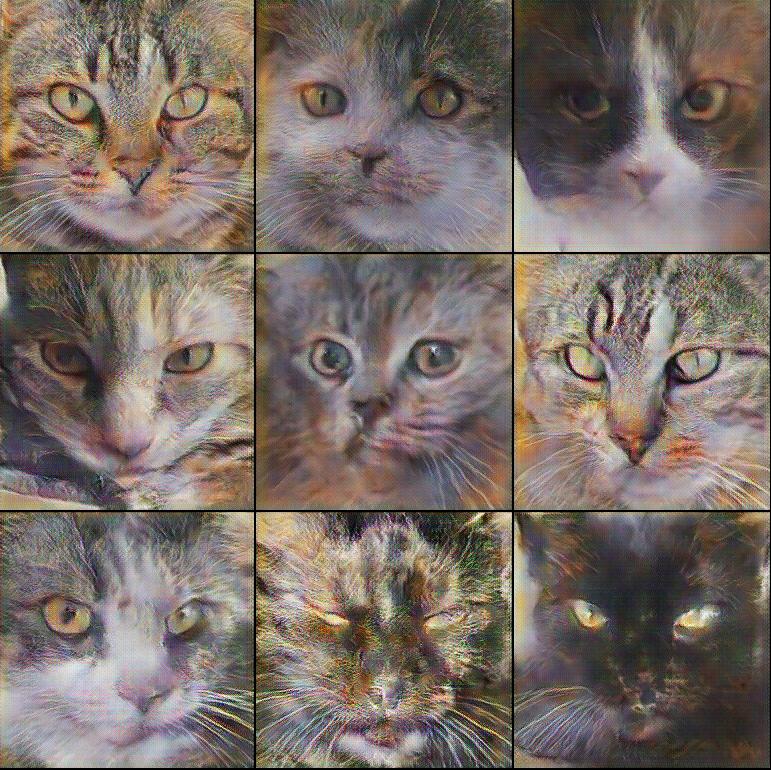}
	\caption{256x256 cats with RaSGAN (FID = 32.11)}
\end{figure}

\begin{figure}[H]
	\centering
	\includegraphics[width=390pt]{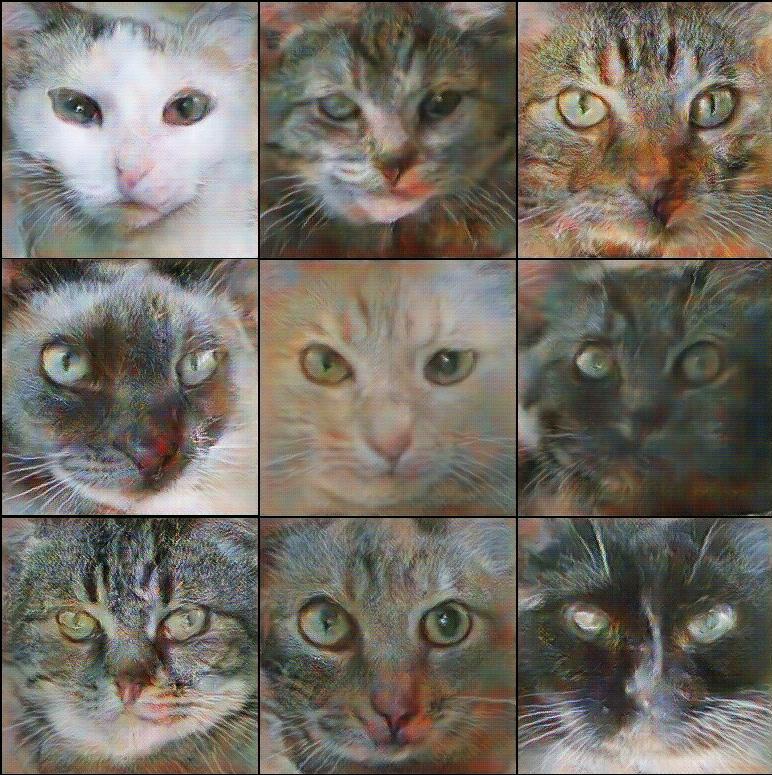}
	\caption{256x256 cats with RaLSGAN (FID = 35.21)}
\end{figure}

\begin{figure}[H]
	\centering
	\includegraphics[width=390pt]{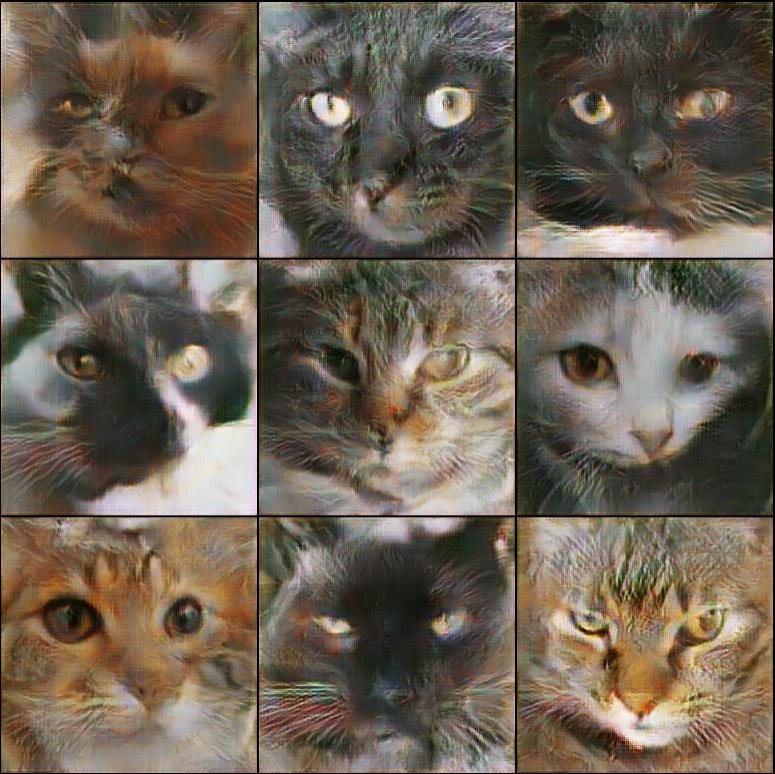}
	\caption{256x256 cats with SpectralSGAN (FID = 54.73)}
\end{figure}

\begin{figure}[H]
	\centering
	\includegraphics[width=390pt]{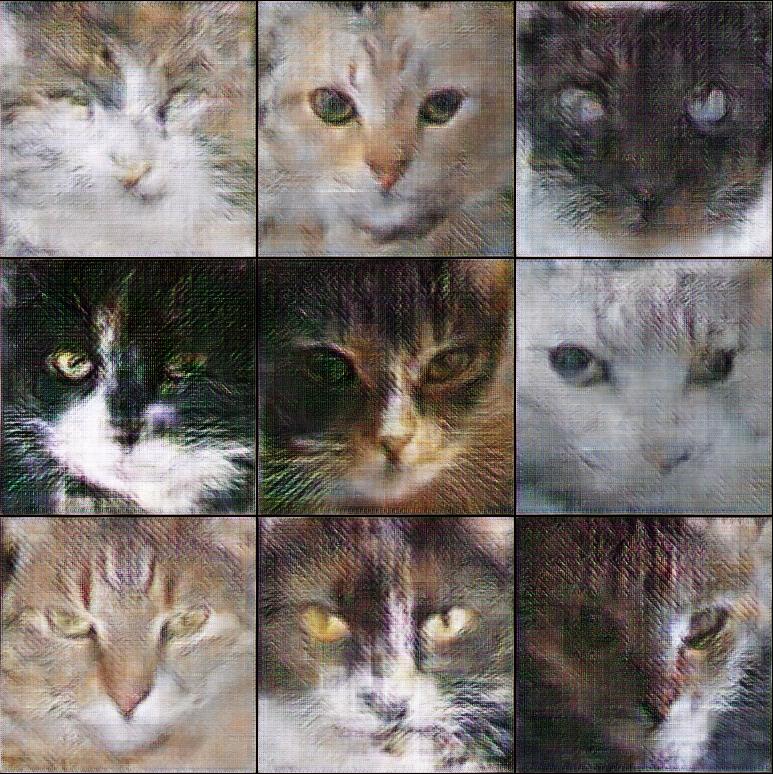}
	\caption{256x256 cats with WGAN-GP (FID > 100)}
\end{figure}

\end{document}